\documentclass[11pt]{article}
\usepackage{blindtext}
\usepackage[utf8]{inputenc}
\usepackage{natbib}
\usepackage{graphicx}
\usepackage{rotating}
\usepackage{pdflscape}
\usepackage{url}
\usepackage{amsmath}
\usepackage{multirow}
\usepackage{lipsum}       
\usepackage{changepage}
\usepackage{enumitem}
\usepackage{caption}
\usepackage{xcolor}
\usepackage{longtable}
\usepackage{hhline}
\usepackage[numbib]{tocbibind}

\title{Neural Relation Prediction for Simple Question Answering over Knowledge Graph}

\author{Amin Abolghasemi, Saeedeh Momtazi\\
Department of Computer Engineering\\
Amirkabir University of Technology
}

\date{}

\begin{document}
\maketitle

\begin{abstract}
\noindent Knowledge graphs are widely used as a typical resource to provide answers to factoid questions. In simple question answering over knowledge graphs, relation extraction aims to predict the relation of a factoid question from a set of predefined relation types. Most recent methods take advantage of neural networks to match a question with all predefined relations. In this paper, we propose an instance-based method to capture the underlying relation of question and to this aim, we detect matching paraphrases of a new question which share the same relation, and their corresponding relation is selected as our prediction. The idea of our model roots in the fact that a relation can be expressed with various forms of questions while these forms share lexically or semantically similar terms and concepts. Our experiments on the SimpleQuestions dataset show that the proposed model achieves better accuracy compared to the state-of-the-art relation extraction models.


\end{abstract}

\section{Introduction}

With the growth of the Internet and rapid production of a vast amount of information, question answering systems, which are designed to find a relevant proper answer by searching throughout a data source, are of great importance.
The production of knowledge bases and the need to answer questions over such resources received researchers’ attentions to propose different models to find the answer of questions from the knowledge bases, known as KBQA\footnote{Knowledge Base Question Answering}. 
Answering factoid questions with one relation, also known as simple question answering, has been widely studied in recent years \citep{Dai:2016,Yin:2016,He:2016,Yu:2017,sawant:2019}. A common approach that has been used in most of the researches is utilizing a two-component system, including an entity linker and a relation extractor. 
In this paper, we focus on the relation extraction component, which is also treated as a classification problem. 

This topic demands certain tools to capture the relation that is mentioned in the questions, as a part of the QA systems. In this paper, we aim to view this kind of relation prediction. The term ``relation extraction'' is originally referred to capturing relation between two entities, if there is any. In the case of simple questions, one of entities is already mentioned and the relation which represents the topic behind the words must be predicted in order to find the other entity. For instance, in the question ``Which artist recorded georgia?'', ``artist'' conveys the topic and "georgia" stands for the first entity. In this context, extracting relation from single-relation questions obtains higher accuracy compared to multi-relational ones. Due to the large number of relations in a knowledge base, however, this simple question relation extraction is not a solved problem yet.

Classifying questions to predefined set of relations is one of the main approaches for this task \citep{Mohammed:2018}. Moreover, matching question content with relations has also been proposed and shown promising results \citep{Yin:2016,Yu:2017}. 
In this paper, the relation extraction is viewed from a new perspective such that relation extraction is done within a question-question matching model, instead of only matching questions and relations. Indeed, while many of the previous works use a matching process between question and relations, we use an instance-based method for classifying relations.
The proposed model benefits from a text matching model, namely MatchPyramid \citep{Pang:2016}, and enhances it with a two-channel model for considering lexical match and semantic match between questions.

The structure of the paper is as follows: in Section \ref{sec:RelatedWorks}, we give a concise overview of the existing approaches for relation classification and its application in KBQA. We also review the available neural text matching models which is the base of our instance-based model. Section \ref{sec:ProposedModel} presents our approach and  elaborately explains detail of our proposed model. In Section \ref{sec:Evaluation}, we show the conducted evaluation experiments and discuss the results. Finally, we summarize our method in Section  \ref{sec:conclusion}.

\section{Related Words}
\label{sec:RelatedWorks}

\subsection{Question Answering over Knowledge Base}

One paradigm in the proposed approaches for relation extraction in KBQA is based on semantic parsing in which questions were parsed and turned into logical forms in order to query the knowledge base \citep{Berant:2013,Berant:2014}. However, most of the recent approaches  \citep{Mohammed:2018,Bordes:2015,Dai:2016,He:2016,Yu:2017} are based on automatically extracted features of terms; thanks to the prominent performance of neural network on representation learning \citep{Mikolov:2013-efficient,Mikolov:2013-distributed}. 

From another point of view, two mainstreams for extracting relations in KBQA are studied: (1) using a classifier which chooses the most probable relation among all  \citep{Mohammed:2018}; (2) matching questions and relations through learning of an embedding space for representing all relations and question words  \citep{Bordes:2015,Dai:2016,Yin:2016,He:2016,Yu:2017}, in which each relation is considered either as a meaningful sequence of words or as a unique entity. \citet{Dai:2016} considered the relation prediction, as well as the whole KBQA problem, as a conditional probability task in which the goal is finding the most probable relation given the question mention. To this aim, they used Gated Recurrent Unit (GRU) neural network and Bidirectional Long Short-Term Memory (BiLSTM) alongside a Conditional Random Field (CRF) for parameterizing their probabilistic component. \citet{Petrochuk:2018} also models the most likely relation distribution using LSTM network.
\citet{He:2016} applied attentional character-level LSTM decoder to embed questions and character-level Convolutional Neural Networks (CNN) to embed knowledge base relations. \citet{Huang:2019} also employs the notion of knowledge graph embedding and learns the relation embeddings using BiLSTM neural network augmented with attention layer . \citet{Yin:2016} applies an attentive max-pooling CNN for matching a question with all relations. Using this attentive max-pooling caused the model to have an augmented representation of question for question-relation matching.

Following \citet{Yin:2016}, \citet{Yu:2017} proposed a hierarchical residual BiLSTM for relation prediction. They used the idea behind residual networks \citep{He:2015} and applied a residual connection to ease the learning process of two layer BiLSTM.
In this research, following \citet{Yu:2017}, we propose a new relation prediction model which uses question-relation matching as well as question-question relevance computation.

\subsection{Neural Text Matching}

The growing area of text matching develops models to investigate the relationship and the degree of matching between sequences of words. This comparison mechanism is a substantial core for various tasks, including ad-hoc retrieval, paraphrase identification, question answering, and semantic web search \citep{Hu:2014}. In this regard, three main categories are considered in the context of deep matching models, namely representation-focused, interaction-focused, and hybrid \citep{Guo:2016}. In representation-focused models, an abstract contextual representation of texts are extracted through neural networks and then these representations are used to estimate the matching score between them. BiMPM \citep{Wang:2017} and ARC I \citep{Hu:2014} are examples of these models. On the other hand, interaction-focused models compute the similarity between two sequences of words in a procedure, such that different patterns and structures of interactions are learned with the help of neural networks based on local interactions of two sequences \citep{Guo:2016}. MatchPyramid \citep{Pang:2016} and aNMM \citep{Yang:2016} are examples of these models. Hybrid models aim to benefit from the advantages of both techniques. ARC II \citep{Hu:2014} is an example of this category.

In this paper, owing to its superior performance, which is reported in Section \ref{sec:Evaluation} , we take advantage of an interaction-focused model in the hierarchy of our model, based on MatchPyramid.

\section{Proposed Model}
\label{sec:ProposedModel}

\begin{figure}[!h]
\centering
\includegraphics[scale=0.4]{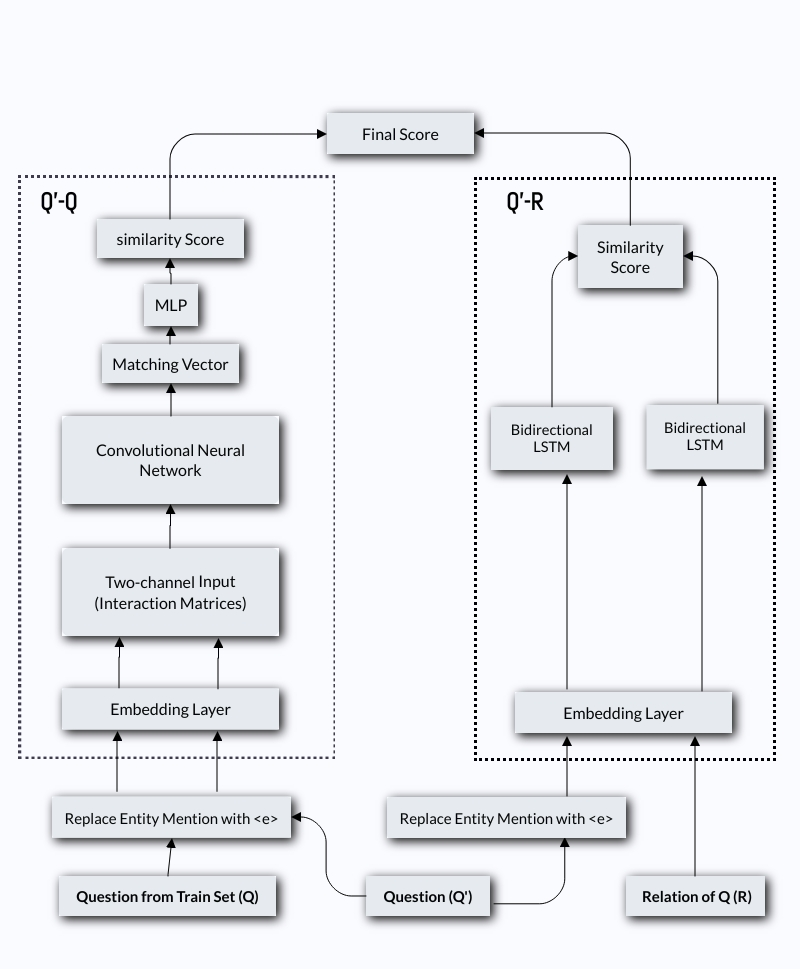}
\captionsetup{font=small}
\caption{Proposed Architecture; including two parallel networks ($Q^{'}-R$ and $Q^{'}-Q$).}
\label{fig:arc}
\end{figure}
In our research, we look at the problem of relation extraction of KBQA from a new point of view to propose our instance-based solution for the task. Before describing the model in detail, we provide an overview of the problem. Given pairs of question and relation in our training data, denoted as ($Q,R$), and pairs of question and relation in our test data, denoted as ($Q^{'}-R^{'}$), for each $Q^{'}$, we aim to predict the most probable relation ($R^{''}$), which interprets the question precisely. Having different lexical representation for each question about a relation,  there are similar words from different range of similarity that occur in the questions of the same relation. Based on these similarities, we argue that the resemblance of questions can be used to detect the relation that lies behind question words. In this regard, following \citet{Yin:2016}, we first extract the entity mentions out of question words and put a symbol (e.g. $<e>$) in its place, so that we will have a question pool in which each question is labeled with its relation that can be considered as a paraphrase of that question. For each new question ($Q^{'}$), we find the most resemble question ($Q$) in our question pool, and assign its corresponding relation as the relation for $Q^{'}$. To do so, we need a model to compute the resemblance of each pair of questions and find the most similar one to $Q^{'}$. Nevertheless, due to existence of multiple form of questions for a relation to be paraphrased, we take the relation of the majority among k top ranked similar questions to $Q^{'}$. In this sense, we are using an instance-based method by computing the relatedness of each new $Q^{'}$ to all train questions.

The architecture of our model is depicted in Figure \ref{fig:arc}. As can be seen, the proposed model consists of two neural networks, namely $Q^{'}-Q$ and $Q^{'}-R$, which work alongside each other and do the computations in parallel in order to provide the output of the whole component. The core idea behind $Q^{'}-Q$ model, the left part of the architecture, is to compute a matching score that represents the similarity of two questions ($Q^{'}$ with Q). Additionally, following \citet{Yu:2017}, we add another neural network ($Q^{'}-R$), the right part of the architecture, to compute the matching score of $Q^{'}$ with the relation of $Q$ ($R$). By doing so, we are enhancing the matching signals between $Q^{'}$ and $Q$ to estimate the overall score.

In the first step, our proposed model projects the input question as well as the available questions and relations of training data into an embedding space. To do so, each sequence of words ($Q^{'}$, $Q$ and $R$) are fed to  an embedding layer and all of their corresponding vectors are fetched. It is noteworthy that, to overcome the problem of unseen relation, each relation is considered as a sequence of words. 
In this work, we use pre-trained vectors due to the fact that current neural embeddings which are learned by using large scale text corpora provide rich enough representation. These embeddings can be enhanced through applying additional learning procedures over knowledge graph \citep{shalaby:2019}. We, however, do not apply these techniques to make sure that the results are comparable with previous works. 
In the next step, input vectors are fed to the two neural networks, which will be described in Sections \ref{sec:Q-Q} and \ref{sec:Q-R}. The output of these two networks are finally combined to produce the final results, described in Section \ref{sec:combination}.

\subsection{$Q^{'}-Q$ Network}
\label{sec:Q-Q}

Both $Q^{'}$ and $Q$ are fed to the $Q^{'}-Q$ network, which is inspired by Matchpyramid \citep{Pang:2016}. Initially, an interaction matrix between the words of $Q^{'}$ and $Q$ is computed.  This interaction matrix has been used in several interaction-based neural learning to match models \citep{Pang:2016,Mitra:2017,Wan:2016,Xiong:2017,Hu:2014},  due to the fact that it provides good representation to compute matching degree between two piece of texts. Indeed, an interaction matrix is a matrix computed based on two sequence of words, in which, each element at the $i^{th}$ row and $j^{th}$ column stands for the similarity between the $i^{th}$ word of the first sequence and $j^{th}$ word of the second sequence. Accordingly, the $(i,j)$ element not only records the exact matches between the words of two sequences, but also estimates the degree of semantic similarity between them \citep{Pang:2016}. 
There are several similarity functions that can be used for creating this matrix; e.g. Cosine similarity, indicator function, dot product, tensor network, etc. . 

In this work, our matrix is computed based on the Cosine similarity. Considering the embedding of $i^{th}$  word of $Q$ as $q_i$ and that of the $j^{th}$ word of $Q^{'}$ as ${q^{'}}_j$ , the $(i,j)$ element of interaction matrix $M$ is represented with the following equation:
\begin{align}
{M_{ij}}^{Cosine} = Cosine(q_i, {q^{'}}_j) = \frac{q_i \cdot{{q^{'}}_j}}{||q_i|| \cdot||{{q^{'}}_j}||}
\end{align}

In addition, we add another interaction matrix in which indicator function is used as the similarity function between $q_i$ and ${q^{'}}_j$:
\begin{align}
{M_{ij}}^{Indicator} = Indicator(q_i, {q^{'}}_j) = \begin{cases}
    1, & \text{if $q_i={q^{'}}_j$}.\\
    0, & \text{otherwise}.
  \end{cases}
\end{align}

The idea behind using indicator function in interaction matrix comes from the fact that the range of difference between words that can be used to express a question about a unique topic/relation (e.g. location of birth) is not too wide and they have relatively low diversity.
By adding indicator matrix to our model, we add an additional bias, which we believe that exists inherently in the problem. This makes our convolutional neural network a 2-channel network, which reflects both lexical and semantic match between text sequences. The impact of adding this extra input channel is discussed  in Section \ref{sec:Evaluation}.
Then, these matching matrices are fed to a convolutional layer to compute a vector (matching vector). The convolution layer operates with a set of kernels $w^{1:n}$ with square sizes of $s_{1:n}$ on the input metrices. The $t^{th}$ kernel slides over the input $M$ and produces feature $v_{ij}^{t}$ as follows:
\begin{align}
v_{ij}^{t} = f(\sum_{c=1}^{S_{t}-1}\sum_{d=1}^{S_{t}-1}{w_{cd}^{t}.M_{i+c,j+d}}  + b^{t})
\end{align}
Here, $f$ represents a non-linear function and $b$ stands for the bias term. The output of the convolution layer $V$ is then passed to a Multi Layer Perceptron (MLP)  to compute the matching score $S_{1}$ between $Q^{'}$ and $Q$ from the question pool. Here, $W_1$ and $W_2$ are the weights of the MLP layer, $b_{i}$ is the bias term, and $o$ is the activation function for the output of $Q^{'}-Q$ network.

\begin{align}
S_{1} =o(W_2f(W_{1}.V + b_{1})+b_{2})
\end{align}

\subsection{$Q^{'}-R$ Network}
\label{sec:Q-R}

Following \citet{Yu:2017}, on the other side of our model, we utilize  a $Q^{'}-R$ network which consists of BiLSTMs to extract contextual representation of question and relation words. Indeed, we are treating the relation words as a meaningful series of words. Then, the computation of similarity between these two enhanced representation is used to estimate the degree of matching between the question $Q^{'}$ and R, which is the corresponding relation of $Q$. Given the computed representation of $Q^{'}$, called $h^{Q^{'}}$, and $R$, called $h^{R}$, the score $S_2$ is represented as follows:  
\begin{align}
S_{2} =Cosine(h^{Q^{'}},h^{R})
\end{align}

\subsection{Combining the Models}
\label{sec:combination}

Finally, overall degree of matching between ($Q,R$) and $Q^{'}$, called $F_{score}$, will be computed. To this aim, different approaches for combination of neural networks are proposed in the literature \citep{Hashem:1997}. In this paper, the final score is computed as a weighted estimation of the scores from both networks while the weights are learned through the learning procedure.

\section{Evaluation}
\label{sec:Evaluation}

\subsection{Dataset}
\label{sec:Dataset}

Following the previous works by \citet{Yin:2016} and \citet{Yu:2017}, we use the common benchmark dataset of the simple question answering, namely SimpleQuestions, which was originally introduced by \citet{Bordes:2015}. This dataset contains 108442 questions gathered with the help of English-speaking annotators. \citet{Yin:2016} proposed a new benchmark for evaluating  relation extraction task on SimpleQuestion. In this benchmark, every question, whose entity  is replaced by a unique token, is labeled with its ground truth relation as its positive label, and all other relation of the gold entity that is mentioned in the question are considered as negative labels. We use the same dataset which contains  72239, 10310 and 20610 question samples as train, validation, and test sets respectively.  To create the data to be used for our training procedure, we choose each relation $R$ and create its question pool $Q^R$. For each sample in the train set which consists a question, its positive (gold) relation, and its negative relations, we select $k_{positive}$ questions from the question pool of the positive relation  $Q^{R^{Positive}}$ and $k_{negative}$ questions from the pool of the negative questions $Q^{R^{Negative}}$. Additionally, for each relation that does not have corresponding questions in the train set, we use their relation words as their questions, and also randomly choose their negative questions from all questions.

\subsection{Experimental Setup}
\label{sec:ExperimentalSetup}

The hyper-parametes are tuned over validation set and they are finally configured as follows: (1) single layer for BiLSTMs (2) 64 units for each BiLSTM, (3) single layer for CNN, and (4) 256 neurons for  the matching vector. The embeddings are initialized with 300-dimension GloVe word vectors \citep{Pennington:2014}. The batch size is fixed to 64 and Adagrad method \citep{Duchi:2011} is used for the optimization.
All the experiments are implemented with Keras library and performed on a Linux machine which has an Intel TMCore i7-6700 3.40 GHz CPU with 16 Gigabyte memory alongside Nvidia GeForce GTX 1080Ti GPU.

\subsection{Results}
\label{sec:Results}

As mentioned, our instance-based idea for question answering over knowledge graph requires a text matching model to find similar question to the input question.

Considering the available text matching models, in the first step of our experiments, we trained different text matching models on the training data. The obtained results are reported in Table \ref{tab:text_matching_res}. As can be seen, the MatchPyramid model performs the best on the proposed model. Considering these results, MatchPyramid has been used as the base of our model in further steps. 

\begin{table}
\begin{center}
\captionsetup{font=small}
\caption{Comparison of different neural text matching models on the proposed instance-based $Q^{'}-Q$ network}
\begin{tabular}{l c}
\hline
Model & Accuracy (\%) \\ \hline
ARC I \citep{Hu:2014} & 89.35\\
ARC II \citep{Hu:2014} & 88.44\\
MatchPyramid \citep{Pang:2016} & 91.75 \\
BiMPM \citep{Wang:2017} & 90.73 \\
\hline
\end{tabular}
\label{tab:text_matching_res}
\end{center}
\end{table}

In the next step, we evaluate the performance of our model while considering exact lexical match as a separate channel in the our $Q^{'}-Q$ network. To this end, three experiments were done and compared: (1) single-channel semantic text matching (Cosine function), (2) single-channel lexical match (indicator function), and (3) two-channel lexical and semantic text matching. The results of these experiments are reported in Table \ref{tab:channel_res}. As can be seen in the results, the impact of adding an extra input channel is obvious as it is compared with single one. The better performance of the semantic channel shows the importance of semantic text matching in the QA system. The further improvement using an additional channel for lexical match indicates that although lexical match is considered implicitly in the normal MatchPyramid model, it is not enough for considering this issue in the QA task and the proposed two-channel model can better cover both semantic and lexical similarities.

\begin{table}
\begin{center}
\captionsetup{font=small}
\caption{The impact of different text matching channels on the proposed instance-based $Q^{'}-Q$ network}
\begin{tabular}{l c}
\hline
Model & Accuracy (\%)\\ \hline
Single-channel semantic text matching & 91.75 \\
Single-channel lexical match & 90.71 \\
Two-channel lexical and semantic text matching & 92.30\\
\hline
\end{tabular}
\label{tab:channel_res}
\end{center}
\end{table}

In the next step of our experiments, we added the $Q^{'}-R$ network to the $Q^{'}-Q$ network and evaluated the new combined architecture, presented in Figure \ref{fig:arc}, on the same dataset. Table \ref{tab:res} reports the performance of our model on classifying the relations in comparison with the state-of-the-art models. 
In this table, AMPCNN \citep{Yin:2016} is an attentive max-pooling CNN for matching a question with all relations. 
APCNN \citep{Santos:2016} and ABCNN \citep{Yin:2016}  both employ an attentive pooling mechanism. These two models are not originally evaluated on relation prediction of simple questions. In fact, the authors of AMPCNN \citep{Yin:2016}, conducted the corresponding experiments on a one-way-attention adaptation of these two models to compare them with the available methods in this task. 
Hier-Res-BiLSTM \citep{Yu:2017} uses hierarchical residual connections to ease the training procedure of BiLSTM.
BiCNN \citep{Yih:2015} uses convolutional neural networks for matching a question with relations. The model is reimplemented for SimpleQuestions by \citet{Yu:2017}.

\begin{table}
\begin{center}
\captionsetup{font=small}
\caption{Results of the proposed model and the state-of-the-art relation prediction models}
\begin{tabular}{l c}
\hline
Model & Accuracy (\%) \\ \hline
AMPCNN \citep{Yin:2016} & 91.3 \\
OWA-APCNN \citep{Santos:2016} &  90.5 \\
OWA-ABCNN \citep{Yin:2016} & 90.2 \\
BiCNN \citep{Yih:2015} &  90.0 \\
Hier-Res-BiLSTM (HR-BiLSTM) \citep{Yu:2017} &  93.3 \\
\hline
Proposed $Q^{'}-Q$ + $Q^{'}-R$ model & 93.41 \\
\hline
\end{tabular}
\label{tab:res}
\end{center}
\end{table}

As it is shown, our proposed model outperforms the state-of-the-art models in relation extraction for SimpleQuestions dataset by a margin of 0.11 percentage. We believe that this improvement is an effect of the two contributions that we had in this paper, namely proposing a combined $Q^{'}-Q$ + $Q^{'}-R$ network and the two-channel text matching model in the $Q^{'}-Q$ network. The combined network helps to consider similarity of questions with other questions in the training data as well as the relations. Adding more matching signals helps to better detect relationship between two questions. More precisely, these signals are from question mentions which are paraphrases of their corresponding relations. This growth in accuracy comes from the aforementioned fact of the inherent low variance of words used in different question forms of an individual relation. 

\subsection{Error Analysis}
In the last step of our experiments, we aim to find the main reasons of errors in the system. To this end, the test questions whose relations were not obtained  correctly in our proposed model are analyzed.

\begin{table}
\begin{center}
\begin{small}
\captionsetup{font=small}
\caption{Examples of errors in the proposed model}
\resizebox{\textwidth}{!}{
\begin{tabular}{l l l}
\hline \hline
Question & Gold Relation & Predicted Relation \\
\hline \hline
what is the genre of the & \texttt{/media\_common/netflix\_title/} & \texttt{/film/film/} \\
 movie $<e>$? & \texttt{netflix\_genres} & \texttt{genre} \\ \hline
is $<e>$ from the united   & \texttt{/people/person/} & \texttt{/people/person/} \\
states or canada? & \texttt{nationality} & \texttt{place\_of\_birth} \\ \hline
what are $<e>$? &  \texttt{/music/album\_content\_type/} & \texttt{/music/genre/} \\
 &  \texttt{/albums} & \texttt{/albums} \\ 
 \hline
what 's an example of & \texttt{/media\_common/literary\_genre/} & \texttt{/book/book\_subject/} \\
a $<e>$ book?  & \texttt{books\_in\_this\_genre} & \texttt{works} \\
\hline  \hline
\end{tabular}
}
\label{tab:errors}
\end{small}
\end{center}
\end{table}

\begin{sloppypar}
Table \ref{tab:errors} presents few examples from those questions.  
Among these questions, there are some predictions in which even the human supervision would assign incorrect relation; e.g., ``what is the genre of the movie $<e>$?'' or ``is $<e>$ from the united states or canada?'', due to very close concepts in the relations or different levels of granularity in the available relations in the knowledge bases. 
In addition, some of questions are practically equivocal; e.g., ``what are $<e>$?'' or ``what's an example of a $<e>$ book?''. Indeed, this ambiguity exists in the training data. Hence, during the training process, an extra variance is imposed to the model. For instance, for the question ``what are $<e>$?'', there are four relations, namely \texttt{/film/film\_genre/films\_in\_this\_genre}, \texttt{/common/topic/notable\_types}, 
and \texttt{/music/album\_content\_type/albums}, \texttt{/cvg/gameplay\_mode/games\_with\_this\_mode} that are assigned to the aforementioned question. It seems that there is an upper bound for relation prediction on SimpleQuestions due to these kinds of indistinctness. 
\end{sloppypar}
\section{Conclusion and Future Work}
\label{sec:conclusion}

In this paper, we proposed a new relation prediction model for simple questions. The proposed model contains two sub-network, a question-question network and a question-relation one, in which we try to match a new sample question with train questions and their corresponding relations respectively. The previous works just employ the semantic matching between a new sample question and relations, whereas our model considered the content of  train questions  while predicting relations. We believe that the words which are used in questions about a relation, convey useful semantic information about that relation. Thus, for future work, we would like to utilize these question words to predict relations from more complex questions.


\bibliographystyle{plainnat}

\bibliography{references}

\begin{thebibliography}{28}
\providecommand{\natexlab}[1]{#1}
\providecommand{\url}[1]{\texttt{#1}}
\expandafter\ifx\csname urlstyle\endcsname\relax
  \providecommand{\doi}[1]{doi: #1}\else
  \providecommand{\doi}{doi: \begingroup \urlstyle{rm}\Url}\fi

\bibitem[Berant and Liang(2014)]{Berant:2014}
Jonathan Berant and Percy Liang.
\newblock Semantic parsing via paraphrasing.
\newblock In \emph{Proceedings of the 52nd Annual Meeting of the Association
  for Computational Linguistics (Volume 1: Long Papers)}, pages 1415--1425,
  Baltimore, Maryland, June 2014. Association for Computational Linguistics.
\newblock \doi{10.3115/v1/P14-1133}.
\newblock URL \url{https://www.aclweb.org/anthology/P14-1133}.

\bibitem[Berant et~al.(2013)Berant, Chou, Frostig, and Liang]{Berant:2013}
Jonathan Berant, Andrew Chou, Roy Frostig, and Percy Liang.
\newblock Semantic parsing on {F}reebase from question-answer pairs.
\newblock In \emph{Proceedings of the 2013 Conference on Empirical Methods in
  Natural Language Processing}, pages 1533--1544, Seattle, Washington, USA,
  October 2013. Association for Computational Linguistics.
\newblock URL \url{https://www.aclweb.org/anthology/D13-1160}.

\bibitem[Bordes et~al.(2015)Bordes, Usunier, Chopra, and Weston]{Bordes:2015}
Antoine Bordes, Nicolas Usunier, Sumit Chopra, and Jason Weston.
\newblock {Large-scale Simple Question Answering with Memory Networks}.
\newblock In \emph{arXiv:1506.02075}, 2015.

\bibitem[Dai et~al.(2016)Dai, Li, and Xu]{Dai:2016}
Zihang Dai, Lei Li, and Wei Xu.
\newblock {CFO}: Conditional focused neural question answering with large-scale
  knowledge bases.
\newblock In \emph{Proceedings of the 54th Annual Meeting of the Association
  for Computational Linguistics (Volume 1: Long Papers)}, pages 800--810,
  Berlin, Germany, August 2016. Association for Computational Linguistics.
\newblock \doi{10.18653/v1/P16-1076}.
\newblock URL \url{https://www.aclweb.org/anthology/P16-1076}.

\bibitem[dos Santos et~al.(2016)dos Santos, Tan, Xiang, and Zhou]{Santos:2016}
Cicero dos Santos, Ming Tan, Bing Xiang, and Bowen Zhou.
\newblock {Attentive pooling networks}.
\newblock In \emph{arXiv:1602.03609}, 2016.

\bibitem[Duchi et~al.(2011)Duchi, Hazan, and Singer]{Duchi:2011}
John Duchi, Elad Hazan, and Yoram Singer.
\newblock Adaptive subgradient methods for online learning and stochastic
  optimization.
\newblock \emph{Journal of machine learning research}, 12\penalty0
  (Jul):\penalty0 2121--2159, 2011.

\bibitem[Guo et~al.(2016)Guo, Fan, Ai, and Croft]{Guo:2016}
Jiafeng Guo, Yixing Fan, Qingyao Ai, and W~Bruce Croft.
\newblock A deep relevance matching model for ad-hoc retrieval.
\newblock In \emph{Proceedings of the 25th ACM International on Conference on
  Information and Knowledge Management}, pages 55--64. ACM, 2016.

\bibitem[Hashem(1997)]{Hashem:1997}
Sherif Hashem.
\newblock Optimal linear combinations of neural networks.
\newblock \emph{Neural networks}, 10\penalty0 (4):\penalty0 599--614, 1997.

\bibitem[He et~al.(2015)He, Zhang, Ren, and Sun]{He:2015}
Kaiming He, Xiangyu Zhang, Shaoqing Ren, and Jian Sun.
\newblock Deep residual learning for image recognition.
\newblock In \emph{IEEE Conference on Computer Vision and Pattern Recognition
  (CVPR)}, pages 770--778, 2015.

\bibitem[He and Golub(2016)]{He:2016}
Xiaodong He and David Golub.
\newblock Character-level question answering with attention.
\newblock In \emph{Proceedings of the 2016 Conference on Empirical Methods in
  Natural Language Processing}, pages 1598--1607, Austin, Texas, November 2016.
  Association for Computational Linguistics.
\newblock \doi{10.18653/v1/D16-1166}.
\newblock URL \url{https://www.aclweb.org/anthology/D16-1166}.

\bibitem[Hu et~al.(2014)Hu, Lu, Li, and Chen]{Hu:2014}
Baotian Hu, Zhengdong Lu, Hang Li, and Qingcai Chen.
\newblock Convolutional neural network architectures for matching natural
  language sentences.
\newblock In \emph{Proceedings of the 27th International Conference on Neural
  Information Processing Systems - Volume 2}, NIPS'14, pages 2042--2050,
  Cambridge, MA, USA, 2014. MIT Press.
\newblock URL \url{http://dl.acm.org/citation.cfm?id=2969033.2969055}.

\bibitem[Huang et~al.(2019)Huang, Zhang, Li, and Li]{Huang:2019}
Xiao Huang, Jingyuan Zhang, Dingcheng Li, and Ping Li.
\newblock Knowledge graph embedding based question answering.
\newblock In \emph{Proceedings of the Twelfth ACM International Conference on
  Web Search and Data Mining}, pages 105--113, 2019.

\bibitem[Mikolov et~al.(2013{\natexlab{a}})Mikolov, Chen, Corrado, and
  Dean]{Mikolov:2013-efficient}
Tomas Mikolov, Kai Chen, Greg~S. Corrado, and Jeffrey Dean.
\newblock {Efficient Estimation of Word Representations in Vector Space},
  2013{\natexlab{a}}.
\newblock URL \url{http://arxiv.org/abs/1301.3781}.

\bibitem[Mikolov et~al.(2013{\natexlab{b}})Mikolov, Sutskever, Chen, Corrado,
  and Dean]{Mikolov:2013-distributed}
Tomas Mikolov, Ilya Sutskever, Kai Chen, Greg~S Corrado, and Jeff Dean.
\newblock Distributed representations of words and phrases and their
  compositionality.
\newblock In C.~J.~C. Burges, L.~Bottou, M.~Welling, Z.~Ghahramani, and K.~Q.
  Weinberger, editors, \emph{Advances in Neural Information Processing Systems
  26}, pages 3111--3119. Curran Associates, Inc., 2013{\natexlab{b}}.

\bibitem[Mitra et~al.(2017)Mitra, Diaz, and Craswell]{Mitra:2017}
Bhaskar Mitra, Fernando Diaz, and Nick Craswell.
\newblock Learning to match using local and distributed representations of text
  for web search.
\newblock In \emph{Proceedings of the 26th International Conference on World
  Wide Web}, WWW '17, pages 1291--1299, Republic and Canton of Geneva,
  Switzerland, 2017. International World Wide Web Conferences Steering
  Committee.
\newblock ISBN 978-1-4503-4913-0.
\newblock \doi{10.1145/3038912.3052579}.
\newblock URL \url{https://doi.org/10.1145/3038912.3052579}.

\bibitem[Mohammed et~al.(2018)Mohammed, Shi, and Lin]{Mohammed:2018}
Salman Mohammed, Peng Shi, and Jimmy Lin.
\newblock Strong baselines for simple question answering over knowledge graphs
  with and without neural networks.
\newblock In \emph{Proceedings of the 2018 Conference of the North {A}merican
  Chapter of the Association for Computational Linguistics: Human Language
  Technologies, Volume 2 (Short Papers)}, pages 291--296, New Orleans,
  Louisiana, June 2018. Association for Computational Linguistics.
\newblock \doi{10.18653/v1/N18-2047}.
\newblock URL \url{https://www.aclweb.org/anthology/N18-2047}.

\bibitem[Pang et~al.(2016)Pang, Lan, Guo, Xu, Wan, and Cheng]{Pang:2016}
Liang Pang, Yanyan Lan, Jiafeng Guo, Jun Xu, Shengxian Wan, and Xueqi Cheng.
\newblock Text matching as image recognition.
\newblock In \emph{Proceedings of the Thirtieth AAAI Conference on Artificial
  Intelligence}, AAAI'16, pages 2793--2799. AAAI Press, 2016.
\newblock URL \url{http://dl.acm.org/citation.cfm?id=3016100.3016292}.

\bibitem[Pennington et~al.(2014)Pennington, Socher, and
  Manning]{Pennington:2014}
Jeffrey Pennington, Richard Socher, and Christopher~D. Manning.
\newblock Glove: Global vectors for word representation.
\newblock In \emph{Empirical Methods in Natural Language Processing (EMNLP)},
  pages 1532--1543, 2014.
\newblock URL \url{http://www.aclweb.org/anthology/D14-1162}.

\bibitem[Petrochuk and Zettlemoyer(2018)]{Petrochuk:2018}
Michael Petrochuk and Luke Zettlemoyer.
\newblock {S}imple{Q}uestions nearly solved: A new upperbound and baseline
  approach.
\newblock In \emph{Proceedings of the 2018 Conference on Empirical Methods in
  Natural Language Processing}, pages 554--558, Brussels, Belgium,
  October-November 2018. Association for Computational Linguistics.
\newblock \doi{10.18653/v1/D18-1051}.
\newblock URL \url{https://www.aclweb.org/anthology/D18-1051}.

\bibitem[Sawant et~al.(2019)Sawant, Garg, Chakrabarti, and
  Ramakrishnan]{sawant:2019}
Uma Sawant, Saurabh Garg, Soumen Chakrabarti, and Ganesh Ramakrishnan.
\newblock Neural architecture for question answering using a knowledge graph
  and web corpus.
\newblock \emph{Information Retrieval Journal}, 22\penalty0 (3-4):\penalty0
  324--349, 2019.

\bibitem[Shalaby et~al.(2019)Shalaby, Zadrozny, and Jin]{shalaby:2019}
Walid Shalaby, Wlodek Zadrozny, and Hongxia Jin.
\newblock Beyond word embeddings: learning entity and concept representations
  from large scale knowledge bases.
\newblock \emph{Information Retrieval Journal}, 22\penalty0 (6):\penalty0
  525--542, 2019.

\bibitem[Wan et~al.(2016)Wan, Lan, Guo, Xu, Pang, and Cheng]{Wan:2016}
Shengxian Wan, Yanyan Lan, Jiafeng Guo, Jun Xu, Liang Pang, and Xueqi Cheng.
\newblock A deep architecture for semantic matching with multiple positional
  sentence representations.
\newblock In \emph{Proceedings of the Thirtieth AAAI Conference on Artificial
  Intelligence}, AAAI'16, pages 2835--2841. AAAI Press, 2016.
\newblock URL \url{http://dl.acm.org/citation.cfm?id=3016100.3016298}.

\bibitem[Wang et~al.(2017)Wang, Hamza, and Florian]{Wang:2017}
Zhiguo Wang, Wael Hamza, and Radu Florian.
\newblock Bilateral multi-perspective matching for natural language sentences.
\newblock In \emph{Proceedings of the 26th International Joint Conference on
  Artificial Intelligence}, pages 4144--4150. AAAI Press, 2017.

\bibitem[Xiong et~al.(2017)Xiong, Dai, Callan, Liu, and Power]{Xiong:2017}
Chenyan Xiong, Zhuyun Dai, Jamie Callan, Zhiyuan Liu, and Russell Power.
\newblock End-to-end neural ad-hoc ranking with kernel pooling.
\newblock In \emph{Proceedings of the 40th International ACM SIGIR Conference
  on Research and Development in Information Retrieval}, SIGIR '17, pages
  55--64, New York, NY, USA, 2017. ACM.
\newblock ISBN 978-1-4503-5022-8.
\newblock \doi{10.1145/3077136.3080809}.
\newblock URL \url{http://doi.acm.org/10.1145/3077136.3080809}.

\bibitem[Yang et~al.(2016)Yang, Ai, Guo, and Croft]{Yang:2016}
Liu Yang, Qingyao Ai, Jiafeng Guo, and W.~Bruce Croft.
\newblock anmm: Ranking short answer texts with attention-based neural matching
  model.
\newblock In \emph{Proceedings of the 25th ACM International on Conference on
  Information and Knowledge Management}, CIKM '16, pages 287--296, New York,
  NY, USA, 2016. ACM.
\newblock ISBN 978-1-4503-4073-1.
\newblock \doi{10.1145/2983323.2983818}.
\newblock URL \url{http://doi.acm.org/10.1145/2983323.2983818}.

\bibitem[Yih et~al.(2015)Yih, Chang, He, and Gao]{Yih:2015}
Wen-tau Yih, Ming-Wei Chang, Xiaodong He, and Jianfeng Gao.
\newblock Semantic parsing via staged query graph generation: Question
  answering with knowledge base.
\newblock In \emph{Proceedings of the 53rd Annual Meeting of the Association
  for Computational Linguistics and the 7th International Joint Conference on
  Natural Language Processing (Volume 1: Long Papers)}, pages 1321--1331,
  Beijing, China, July 2015. Association for Computational Linguistics.
\newblock \doi{10.3115/v1/P15-1128}.
\newblock URL \url{https://www.aclweb.org/anthology/P15-1128}.

\bibitem[Yin et~al.(2016)Yin, Yu, Xiang, Zhou, and Sch{\"u}tze]{Yin:2016}
Wenpeng Yin, Mo~Yu, Bing Xiang, Bowen Zhou, and Hinrich Sch{\"u}tze.
\newblock Simple question answering by attentive convolutional neural network.
\newblock In \emph{Proceedings of {COLING} 2016, the 26th International
  Conference on Computational Linguistics: Technical Papers}, pages 1746--1756,
  Osaka, Japan, December 2016. The COLING 2016 Organizing Committee.
\newblock URL \url{https://www.aclweb.org/anthology/C16-1164}.

\bibitem[Yu et~al.(2017)Yu, Yin, Hasan, dos Santos, Xiang, and Zhou]{Yu:2017}
Mo~Yu, Wenpeng Yin, Kazi~Saidul Hasan, Cicero dos Santos, Bing Xiang, and Bowen
  Zhou.
\newblock Improved neural relation detection for knowledge base question
  answering.
\newblock In \emph{Proceedings of the 55th Annual Meeting of the Association
  for Computational Linguistics (Volume 1: Long Papers)}, pages 571--581,
  Vancouver, Canada, July 2017. Association for Computational Linguistics.
\newblock \doi{10.18653/v1/P17-1053}.
\newblock URL \url{https://www.aclweb.org/anthology/P17-1053}.

\end{thebibliography}

\end{document}